\documentclass[11pt]{article}
\usepackage[]{EMNLP2022}
\usepackage{times,latexsym}
\usepackage[utf8]{inputenc}
\usepackage[T1]{fontenc}
\usepackage{amsmath}
\usepackage{amssymb}
\usepackage{url}
\usepackage{xcolor}
\usepackage{graphicx}
\usepackage{microtype}
\usepackage{verbatim}
\usepackage{enumitem}
\usepackage{caption}
\usepackage{subcaption}
\usepackage{fixmath}
\usepackage{listings}
\AtBeginDocument{
\addtolength{\abovedisplayskip}{-2ex}
\addtolength{\abovedisplayshortskip}{-2ex}
}

\newcommand{\grad}{\nabla_\theta}
\newcommand{\joint}{\mathcal{L}}
\newcommand{\marginal}{\mathcal{ML}}
\newcommand{\conditional}{\mathcal{CL}}
\newcommand{\pth}{p_{\theta}}
\newcommand{\tightcbox}[2]{\setlength{\fboxsep}{0pt}\colorbox{#1}{\strut #2}}
\newcommand{\htheta}{\hat{\theta}}
\renewcommand{\vec}{\underline}

\title{Influence Functions for Sequence Tagging Models}

\author{Sarthak Jain \\ Northeastern University \\\texttt{jain.sar@northeastern.edu} \\\And
  Varun Manjunatha \\ Adobe Research \\ \texttt{vmanjuna@adobe.com} \\ \AND 
  Byron C. Wallace \\ Northeastern University \\ \texttt{b.wallace@northeastern.edu} \\ \And
  Ani Nenkova \\ Adobe Research \\ \texttt{nenkova@adobe.com}}

\begin{document}

\maketitle

\begin{abstract}

Many language tasks (e.g., Named Entity Recognition, Part-of-Speech tagging, and Semantic Role Labeling) are naturally framed as \emph{sequence tagging} problems. However, there has been comparatively little work on interpretability methods for sequence tagging models. In this paper, we extend \emph{influence functions} --- which aim to trace predictions back to the training points that informed them --- to sequence tagging tasks. We define the influence of a training instance \emph{segment} as the effect that perturbing the labels within this segment has on a test segment level prediction. We provide an efficient approximation to compute this, and show that it tracks with the true segment influence, measured empirically. We show the practical utility of segment influence by using the method to identify systematic annotation errors in two named entity recognition corpora. Code to reproduce our results is available at \url{https://github.com/successar/Segment_Influence_Functions}.
\end{abstract}

\section{Introduction}

Instance attribution methods aim to identify training examples that most informed a particular (test) prediction. The influence of training point $k$ on test point $i$ is typically formalized as the change in loss that would be observed for point $i$ if example $k$ was removed from the training set \cite{koh2017understanding}. Heuristic alternatives have also been developed to measure the importance of training samples during prediction, such as retrieving training examples similar to a test item  
\cite{pezeshkpour2021empirical,ilyas2022datamodels,guo2021fastif}. 

Influence functions can facilitate dataset debugging by helping to surface training samples which exhibit artifacts \cite{han2020explaining}. But on language tasks, most work on identifying training samples influential to a particular prediction has focused on classification tasks \cite{koh2017understanding,han2020explaining,pezeshkpour2021empirical}.
It is not immediately clear how we can extend such methods to the structured prediction problems such as named entity recognition (NER). 

\begin{figure}
    \centering
    \includegraphics[scale=0.6]{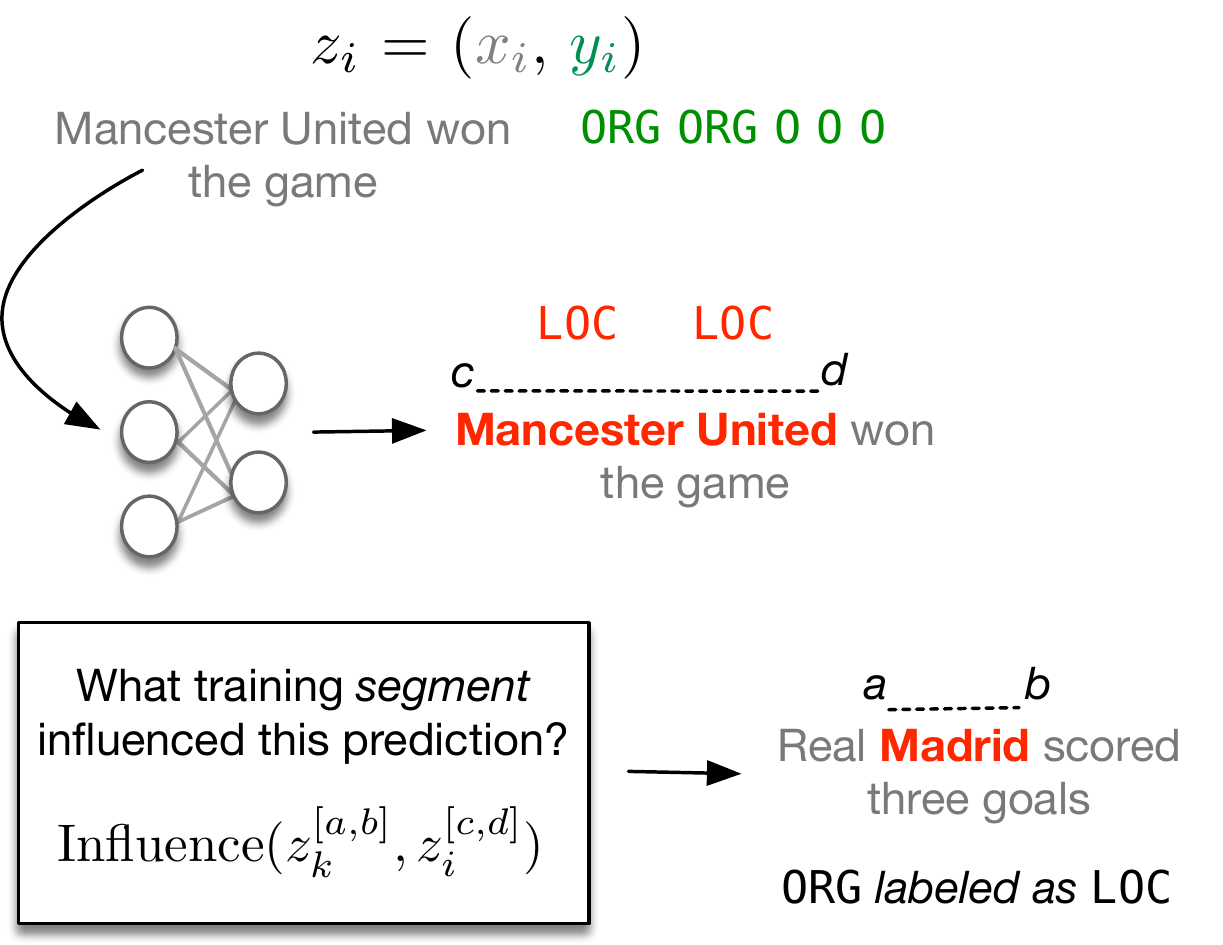}
    \caption{We propose and evaluate \emph{influence functions for sequence tagging tasks}, which retrieve snippets (from token $a$ to $b$) in train samples that most influenced predictions for test tokens $c$ through $d$. Here this reveals a training example in which an {\tt ORG} is problematically marked as a {\tt LOC}, leading to the observed error.}
     \vspace{-.5em}
    \label{fig:seq-inf}
\end{figure}

In this work we address this gap, presenting new methods for characterizing token-level influence for structured predictions (specifically sequence tagging tasks), and evaluating their use across illustrative datasets. More specifically, we focus on NER, one of the most common sequence tagging tasks. We extend influence functions to detect important training examples, i.e., those that most influenced the prediction \emph{of a specific entity}, as opposed to being most influential with respect to the entire predicted label sequence. We call this extension \emph{segment influence}. 

Segment influence can help one perform fine-grained analysis of why specific segments of text were incorrectly labeled (as opposed to the entire sequence). Consider, for example, Figure~\ref{fig:seq-inf}. This shows a common issue in the CoNLL NER dataset: city names contained in soccer club titles tend to be mislabeled as location, rather than organization. 
This in turn leads to similar mispredictions in the test set. For the shown test example we can use segment influence to ask \emph{which entities within the training examples most informed the prediction made for the entity `Manchester United'}? In principal, segment influence can directly recover the entities  responsible for this systematic mislabeling. 

Our main {\bf contributions} are as follows. (1) We present a new method to approximately compute token-level influence for outputs produced by sequence tagging models; (2) We evaluate whether this approximation corresponds to \emph{exact} influence values in linear models, and whether the method recovers intuitively correct training examples in synthetically constructed cases, and; (3) We establish the practical utility of approximating structured influence by using the method to identify systematic annotation errors in NER corpora.

\section{Influence for Sequence Tagging}

Consider a standard \emph{sequence tagging} task in which the aim is to estimate the parameters $\theta$ of function $f_\theta$ which assigns to each token $x_{it} \in \mathcal{V}$ in an input sequence $x_i$ (of length $T_i$) a label $y_{it}$ from a label set $\mathcal{Y}$. Denote the training dataset by $\mathcal{D}$ where $\mathcal{D} = \{(x_i = \{x_{it}\}_{t=1}^{T_i}, y_i = \{y_{it}\}_{t=1}^{T_i})\}$. 

Define $f_{\theta}$ as a model that yields conditional probability estimates for sequence label assignments: $p_{\theta}(y_i | x_i)$. 
Given parameter estimates $\hat{\theta}$, we can make a prediction for a test instance $x_i$ by selecting the most likely $y$ under this model: 
$\hat{y}_i = \text{argmax}_{y} p_{\hat{\theta}}(y | x_i)$. 
In structured prediction tasks we assume that the label $y_{it}$ depends in part on labels $y_i \setminus y_{it}$, given the input $x_i$. 
In linear chain sequence tagging, this dependence can be formalized as a graphical model in which adjacent labels are connected; the most common realization of such a model is perhaps the \emph{Conditional Random Field} (CRF; \citealt{CRF-ICML}).

We typically estimate $\theta$ by minimizing the negative log-likelihood of the training dataset $\mathcal{D}$.
\begin{equation}
\label{model-argmin}
    \underset{\theta}{\arg\!\min}\hspace{.2em} \frac{1}{|\mathcal{D}|} \sum_{(x_i, y_i) \in \mathcal{D}} -\log p_{\theta}(y_i | x_i)
\end{equation}

\noindent For brevity, we will also write the loss (negative log likelihood) of an example $z_i = (x_i, y_i)$ as $\mathcal{L}(z_i, \theta) = -\log p_{\theta}(y_i|x_i)$ and the overall loss over the training set by $\mathcal{L}(\mathcal{D}, \theta) = \frac{1}{|\mathcal{D}|} \sum_{z_i \in \mathcal{D}} \mathcal{L}(z_i, \theta)$.

\subsection{Background: Influence Functions in ML}

\noindent\textbf{Influence Functions} \cite{koh2017understanding} retrieve training samples $z_k$ deemed ``influential'' to the prediction made for a specific test sample $x_i$: $\hat{y}_i = f_{\hat{\theta}}(x_i)$. The exact influence of a training example $z_k$ on a test example $z_i=(x_i, y_i)$ is defined as the change in the loss on $z_i$ that would be incurred under parameter estimates \emph{if the training sample $z_k$ were removed prior to training}, i.e., $\mathcal{L}(z_i, \hat{\theta}_{-z_k}) - \mathcal{L}(z_i, \hat{\theta})$. 

In practice this is prohibitively expensive to compute. \citet{koh2017understanding} proposed an approximation. The idea is to measure the change in the loss on $z_i$ observed when the loss associated with train sample $z_k$ is slightly upweighted by some $\epsilon$. Explicitly computing the effect of such an $\epsilon$-perturbation is not feasible. \citet{koh2017understanding} provide an efficient mechanism to approximate this (reproduced in Appendix~\ref{app:instance-inf-deriv}): $I(z_i, z_k) = -\nabla_{\theta} \mathcal{L}(z_i, \htheta) [\nabla^2_{\theta} \joint(\mathcal{D}, \htheta)]^{-1} \nabla_{\theta} \mathcal{L}(z_k, \htheta)$, where $\nabla^2_{\theta} \joint(\mathcal{D}, \htheta)$ is the Hessian of the loss $\mathcal{L}(\mathcal{D}, \htheta)$ over the dataset with respect to $\theta$.

Sequence tagging tasks by definition involve \emph{multiple} predictions (and labels) per instance, and it is therefore natural to consider finer-grained influence. 
In particular, we would like to quantify the effect of \emph{segments} of labels for $z_k$ on a specific segment of the predicted output for $z_i$. For example, if we mispredict a particular entity within $x_i$, we may want to identify the train sample segment(s) most responsible for this error, especially if the model makes \emph{systematic} errors that might be rectified by cleaning $\mathcal{D}$. 

\subsection{Segment Level Influence} 

We provide machinery to compute segment level influence. 
We want to quantify the impact of training tokens $x_{k[a, b]}$ (with labels $y_{k[a, b]}$), $1 \le a, b \le T_k$ on the loss of a segment of test point $z_i$. In NER, these segments may correspond to entities. 

\subsubsection{Exact Segment Influence}
\label{sec:exact-inf}

We define the \emph{exact} influence of a segment $[a, b]$ within training example $z_k$ on a segment $[c, d]$ of a test example $z_i = (x_i, y_i)$ as the change in loss that would be observed for reference token labels in segment $[c, d]$ of $z_i$, had we excluded the labels for segment $[a, b]$ within $z_k$ from the training data. To make the above definition precise, we need to first define how training is to be performed when only partial annotations may be available for a given train example (i.e., where a segment has been ``removed''). 
We need also to formally define change in loss for a \emph{segment} of a test example.

Start with training under partial annotations. Consider a training example $z_k=(x_k=\{x_{k1}, \dots, x_{kT_k}\}, y_k=\{y_{k1}, \dots, y_{kT_k}\})$. Assume we did not have labels for segment $[a, b]$ in $y_k$, i.e., labels $\{y_{ka}, \dots, y_{kb}\}$ were missing. Denote such a partial label sequence by $y_k^{-[a, b]} = y_k \setminus \{y_{ka}, \dots, y_{kb}\}$. Let $y_k^{[a, b]} = \{y_{ka}, \dots, y_{kb}\}$. 
A natural way to handle such cases is to marginalize over all possible label assignments to the segment $[a, b]$ when computing the likelihood of this training example \cite{tsuboi-etal-2008-training}:

\begin{equation}
\begin{split}
    &p_{\theta}(y_k^{-[a, b]}|x_k) = \\ &\sum_{y'_a \in \mathcal{Y}} \dots \sum_{y'_b \in \mathcal{Y}} p_{\theta}(y_k^{-[a,b]} \cup \{y'_a, ..., y'_b\}|x_k) 
\end{split}
\end{equation}

\noindent Denote the \textbf{marginal loss} of this partially annotated sequence as $\marginal(z_k^{-[a,b]}, \theta) = -\log p_{\theta}(y_k^{-[a, b]}|x_k)$.

We can also write this marginal loss as the difference between the joint loss of $y_k$ and the conditional loss of the segment $y_k^{[a, b]}$. This second form is more intuitive when we move to approximate exact influence values via $\epsilon$-weighting. 

\begin{equation}
\begin{split}
    \log p_{\theta}(y_k^{-[a, b]}&|x_k) = \log p_{\theta}(y_k|x_k) \\&- \log p_{\theta}(y_k^{[a, b]} | y_k^{-[a, b]}, x_k) 
\end{split}
\end{equation}
\begin{equation}
\label{eq:marg-loss-def}
\marginal(z_k^{-[a,b]}, \theta) = \mathcal{L}(z_k, \theta) - \conditional(z_k^{[a, b]}, \theta)
\end{equation}

\noindent where we have defined also the \textbf{conditional loss} of the segment as $\conditional(z_k^{[a, b]}, \theta) = -\log p_{\theta}(y_k^{[a, b]} | y_k^{-[a, b]}, x_k)$.

Next we define the change in loss for a segment of a test example $z_i = (x_i, y_i)$. 
We define the loss for the segment $[c, d]$ of the output $y_i$ as the conditional loss of the segment $[c, d]$: $\conditional(z_i^{[c, d]}, \theta)$.

Given the above definitions, we can  concretize the notion of the \emph{exact} influence as follows:


\vspace{0.35em}
\noindent 1. Retrain the model without the segment $[a, b]$ of training example $z_k$:
    \begin{equation}
    \label{eq:exact-inf-retrain}
    \begin{split}
        \hat{\theta}&[z_k^{-[a, b]}] = \underset{\theta}{\arg\!\min}\hspace{.2em} \frac{1}{|\mathcal{D}|} \sum_{z_l \in \mathcal{\mathcal{D}}} \mathcal{L}(z_l, \theta) \\
        &-\frac{1}{|\mathcal{D}|} (\mathcal{L}(z_k, \theta) - \marginal(z_k^{-[a, b]}, \theta))
    \end{split}
    \end{equation}
    
    \noindent Comparing Equations~\ref{eq:marg-loss-def} and \ref{eq:exact-inf-retrain}, we see that removing the effect of segment $[a,b]$ of $z_k$  amounts to subtracting the conditional loss of the segment $\conditional(z_k^{-[a, b]}, \theta)$ from the original loss $\mathcal{L}(D, \theta)$.
    
    
\vspace{0.35em}
\noindent 2. Compute the difference between the conditional loss of segment $[c, d]$ of test example $z_i$ under new parameter estimates $\htheta[z_k^{-[a, b]}]$ and the original estimates $\htheta$ trained using the objective in Equation~\ref{model-argmin}:
    \begin{equation}
    \label{eq:exact-inf-def}
    \begin{split}
        &\text{Exact-Influence}(z_k^{[a, b]}, z_i^{[c, d]}) = 
        \\ &\conditional(z_i^{[c, d]}, \htheta[z_k^{-[a, b]}]) - \conditional(z_i^{[c, d]}, \htheta)
    \end{split}
    \end{equation}

\subsubsection{Approximating Segment Influence}
\label{sec:approx-inf}

Above we have derived a means to calculate exact segment level influence values. But in practice retraining the model (step 1) is not feasible. Here we instead present an $\epsilon$-upweighting method --- analogous to the approximation to instance influence \cite{koh2017understanding} --- for computing segment influence. The idea is to compute the change in model parameters if we incur a slight additional penalty $\conditional(z_k^{-[a, b]}, \theta)$ for the segment $[a, b]$:

\begin{equation}
\label{eq:segment-upweigh-def}
\begin{split}
    \htheta_{\epsilon}[z_k^{-[a, b]}] &= \underset{\theta}{\arg\!\min}\hspace{.2em} \frac{1}{|D|} \sum_{z_l \in \mathcal{D}} \mathcal{L}(z_l, \theta) \\
    &+ \epsilon \hspace{.3em}\conditional(z_k^{[a, b]}, \theta)
\end{split}
\end{equation}

\noindent A first order approximation to the difference in the model parameters near $\epsilon = 0$ is given by:

\begin{equation}
    \frac{d\htheta_{\epsilon}[z_k^{-[a, b]}]}{d\epsilon}\Bigr|_{\epsilon=0} = -[\nabla^2_{\theta} \joint(\mathcal{D}, \htheta)]^{-1} \nabla_{\theta} \conditional(z_k^{[a, b]}, \htheta)
\end{equation}

\noindent We can apply the chain rule to measure the change in the conditional loss over segment $[c, d]$ of test example $z_i$ due to this upweighting:

\begin{equation}
\begin{split}
    &\text{I}(z_i^{[c,d]}, z_k^{[a, b]}) = \frac{d\hspace{0.2em}\conditional(z_i^{[c, d]}, \htheta_{\epsilon}[z_k^{-[a, b]}])}{d\epsilon}\Bigr|_{\epsilon=0} =\\
    &-\nabla_{\theta} \conditional(z_i^{[c, d]}, \htheta) [\nabla^2_{\theta} \joint(\mathcal{D}, \htheta)]^{-1} \nabla_{\theta} \conditional(z_k^{[a, b]}, \htheta)
\end{split}
\label{eq:segment-influence-approx}
\end{equation}

This definition provides us with an approximation to the exact influence defined in previous section for a segment of a training example on a segment of a test example. Derivation showing all the steps can be found in Appendix~\ref{app:segment-inf-deriv}.
We have assumed that we can take the gradient of the conditional loss over a segment, which  
is possible for a CRF tagger (derivation in  Appendix~\ref{app:crf-grad}), but may be non-trivial for other models.

\section{Computational Challenges}
\label{sec:challenges}
\vspace{-.25em}
The computation costs of even approximate instance level influence can be prohibitive in practice, especially with the large pretrained language models that now dominate NLP. 
Computing and storing inverse Hessians of the loss has $O(p^3)$ and $O(p^2)$ complexity where $p$ is the number of parameters in the model (commonly $\sim$100M-100B for deep models). Even ignoring the Hessian, one still needs the gradient with respect to each training example in $\mathcal{D}$; one could attempt to pre-compute these, but storing the results requires $O(|D|p)$ memory. 

The alternative is therefore to recompute these for each new test point $z_i$. 
For segment level influence these costs are compounded because we need influence with respect to \emph{every} segment within a training example, multiplying complexities by $T^2$, where $T$ is the average length of a training example. 
Consequently, it is practically infeasible to calculate segment influence per Equation \ref{eq:segment-influence-approx}.

Prior work by \citet{pezeshkpour2021empirical} showed that for instance-level influence, considering a restricted set of parameters (e.g., those in the classification layer) when taking the gradient is reasonable for influence in that this does not much affect the induced rankings over train points with respect to influence.  
Similarly, ignoring the Hessian term 
does not significantly affect rankings by influence. 
These two simplifications dramatically improve efficiency, and we adopt them here.

Consider a sequence tagging model built on top of a deep encoder $F$, e.g., BERT \cite{devlin2018bert}. 
In the context of a linear chain CRF on top of $F$, the standard score function for this model is: $s(y_i, x_i) = \sum_{t=1}^{T_i} \vec{y}_{it}^\top \mathbold{W} F(x_i)_t + \vec{y}_{i(t-1)}^\top \mathbold{T} \vec{y}_{it}$, where $\mathbold{T}$ is a matrix of class transition scores and $\vec{y}_{it}$ is the one-hot representation of label $y_{it}$. 
A CRF layer consumes these scores and computes the probability of a label sequence as $p(y_i|x_i) = \frac{e^{s(y_i, x_i)}}{\sum_{y' \in \mathcal{Y}^{T_i}} e^{s(y', x_i)}}$. In this work we consider the gradient only with respect to the $\mathbold{W}$ and $\mathbold{T}$ parameters above and not any parameters associated with $F$. 

Further, we consider influence only with respect to individual token outputs in training samples, rather than every possible segment --- i.e., we only consider single-token segments. This further reduces the $T^2$ terms in our complexity to $T$.

\section{Experimental Aims and Setup}
\label{sec:data-model}
We evaluate segment influence in terms of: (1) \emph{Approximation validity}, or how well the approximation proposed in Equation~\ref{eq:segment-influence-approx} corresponds to the exact influence value  (Equation~\ref{eq:exact-inf-def}), and, (2) \emph{Utility}, i.e., whether segment influence might help identify problematic training examples for sequence tagging tasks. 
In this paper we consider only NER tasks, using the following datasets.

\noindent{\bf CoNLL} \cite{tjong-kim-sang-de-meulder-2003-introduction}. An NER dataset containing news stories from Reuters, labeled with four entity types: {\tt PER}, {\tt LOC}, {\tt ORG} and {\tt MISC}, and demarcated using the beginning-inside-out (BIO) tagging scheme \cite{ramshaw1999text}. The dataset is divided into train, validation, and test splits comprising 879, 194, and 197 documents, respectively.

\noindent{\bf EBM-NLP} \cite{nye-etal-2018-corpus}. 
A corpus of annotated medical article abstracts describing clinical randomized controlled trials. 
`Entities' here are spans of tokens that describe the patient Population enrolled, the Interventions compared, and the Outcomes measured (`PICO' elements). 
This dataset includes a test set of 191 abstracts labeled by medical professionals, and a train/val set of 3836/958 abstracts labeled via Amazon Mechanical Turk. 

For NER models we use a representative modern transformer --- BigBird Base ~\cite{zaheer2020big} --- as our encoder $F$, using final layer hidden states as contextualized token representations. 
Dependencies between output labels are modeled using a CRF. We provide training details in Appendix~\ref{app:data-model}.

In addition to segment and instance level influence we also evaluate --- where applicable --- {\bf segment nearest neighbor} as an attribution method, which works as follows. 
We retrieve from the training dataset segments that have the highest similarity between their corresponding feature embeddings (we again consider only single token segments here, so we do not need to worry about embedding multi-token segments). We consider both dot product and cosine similarity and report the results for the version that gives best performance for each experiment. 

\section{Validating Segment Influence}
\vspace{-.25em}
In this first set of experiments we aim to (i) verify that the proposed approximation correlates with the \emph{exact} influence (\ref{sec:approx-inf}), and, (ii) ascertain that for synthetic tasks which we construct, segment influence returns the \emph{a priori} expected training segment for given text examples (\ref{sec:synthetic}).
In the latter experiments, we compare the fine-grained error analysis afforded by segment influence to that made possible using instance-level attribution.

\subsection{How Good is our Approximation?}
\label{sec:corr}

How well does the \emph{approximation} we have proposed for segment influence correlate with the \emph{exact} value? The latter is in general intractable to compute. As a practical means to validate our approximation we use a simple linear model trained on the CoNLL corpus, which makes it feasible (though still costly) to compute exact influence via brute force. This allows us to compare the actual segment influence to our approximation. 

We subset our training set to a 1000 examples and the validation set to 200 examples (given the computational expense of model retraining). As token-level features, we use GloVe word embeddings \cite{pennington2014glove} and additional syntactical features (See Appendix~\ref{app:syntac-feats}). We use the L-BFGS optimizer available in {\tt PyTorch} \cite{NEURIPS2019_9015} to train this model, stopping when the maximum absolute gradient value reaches $10^{-6}$. 

We randomly sample 20 mispredicted validation tokens. For each such token, we identify the 20 most influential tokens in the training set according to their absolute approximated influence values. We combine these 400 tokens together in a single pool, remove these tokens sequentially prior to training by subtracting their conditional loss (Equation~\ref{eq:exact-inf-def}), and retrain the model. We then take the difference in the observed loss for each of the 20 validation token under the retrained and original model.

\begin{figure}
    \centering
    \includegraphics[scale=0.425]{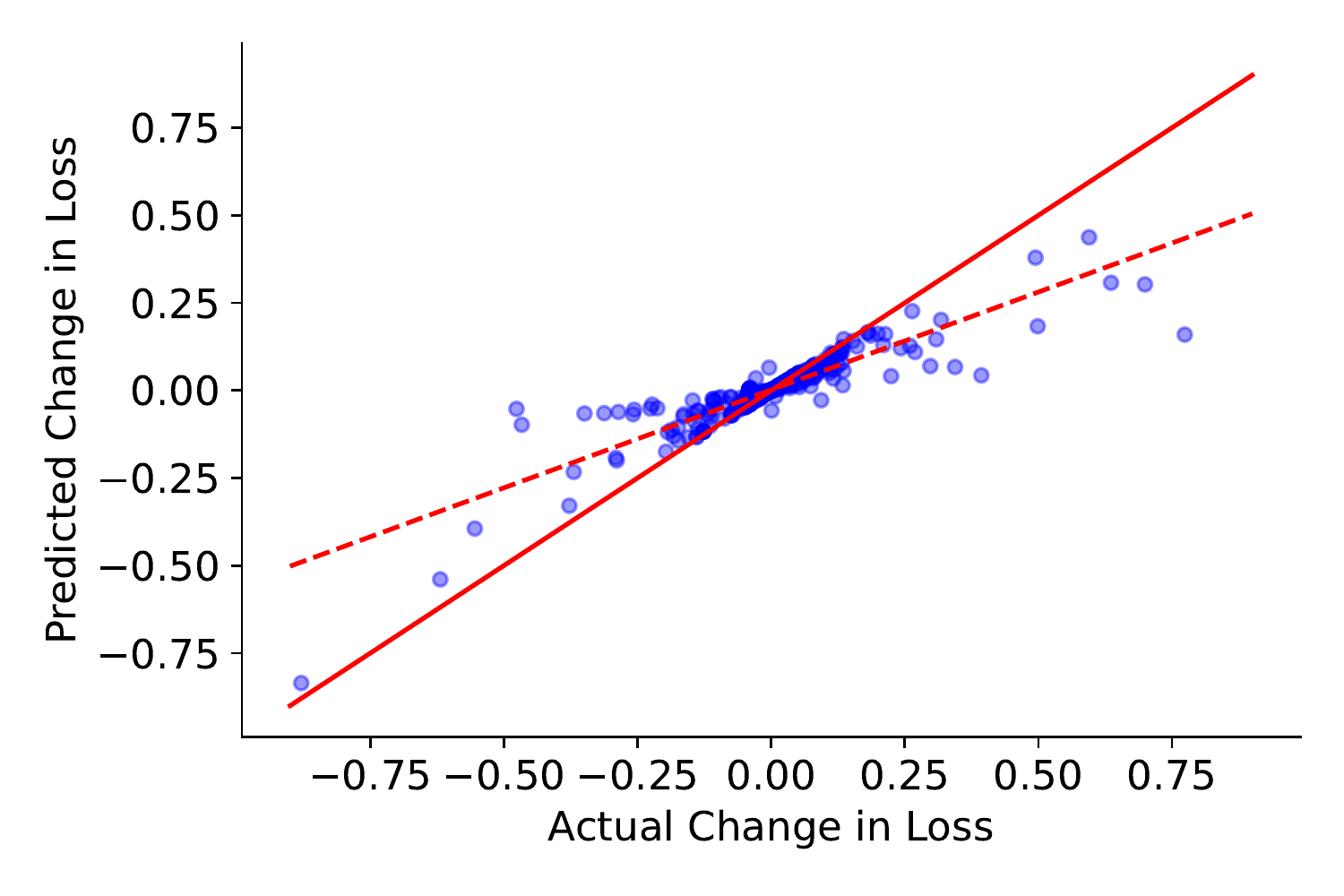}
    \vspace{-.65em}
    \caption{Predicted versus actual change in conditional loss over a sample of mispredicted validation tokens from CoNLL. The dotted line shows the line of best fit to the observations (solid line is perfect correlation).}
    \vspace{-.75em}
    \label{fig:exact-vs-approx}
\end{figure}

Figure~\ref{fig:exact-vs-approx} plots the \emph{actual} difference in the conditional loss obtained for the sampled validation tokens against the \emph{predicted} change in loss using the influence approximation (Equation~\ref{eq:segment-influence-approx}). 
The quantities have a Pearson correlation of 0.89, suggesting that the approximation is reasonable, though imperfect; some deviations exist (likely due to numerical instability computing the inverse Hessian).

\subsection{Synthetic Artifact Insertion}
\label{sec:synthetic}

We next evaluate whether segment influence may help practitioners `debug' dataset issues by identifying (problematic) entities within train examples that may have led to test time entity mispredictions.  

We again use the CoNLL training set for this exercise, artificially inserting an ``artifact'' into training samples. Specifically, for a random 10\% of train instances, we insert an ``artifact token'' ({\tt special}) at a position selected at uniform random from range $1$ to $T$-1. When this insertion is made, we also deterministically set the label of the token immediately following the artifact to {\tt B-PER}. As an example, consider the example \emph{Joe}/{\tt PER} \emph{Biden}/{\tt PER} \emph{is US}/{\tt LOC} \emph{president in 2021}. We might modify this to the following: \emph{Joe}/{\tt PER} \emph{Biden}/{\tt PER} \emph{is US}/{\tt LOC} \emph{president} {\tt special} \emph{in}/{\tt PER} \emph{2021}. We train a model on this modified train set. 

If the model learns the correlation between artifact tokens and the entity label for the right adjacent token, then for test examples featuring the artifact it should return artifact adjacent tokens (e.g., \emph{in} in the above example) as influential. 
To test this, we insert the artifact randomly into 10\% of the validation instances. 
The model predicts {\tt PER} for all tokens that immediately follow the artifact. 
We apply segment influence for these token predictions, and find that for \emph{all} such examples, 
the most influential token retrieved using segment influence is adjacent to the inserted artifact. Segment nearest neighbor also recovers these problematic samples. 
By contrast, applying instance influence to the same validation examples yields training examples exhibiting the artifact as most influential only 26.3\% of the time.

This result provides evidence that we are able to retrieve plausible error-causing entities within the training dataset that affect a test entity prediction. Standard instance-level influence is not as useful here because it evaluates influence with respect to an entire example, and not in terms of its constituent entities. This highlights the need for and potential utility of \emph{segment} influence methods. 

\section{Use Cases for Segment Influence}
\vspace{-.3em} 
Misannotations or incomplete annotations are a common problem in sequence tagging tasks. 
In contrast to standard classification tasks, in the case of sequence tagging annotators must label multiple spans of variable length within a given text. The decision of where exactly to begin and end such spans is often inherently difficult, and so can result in label inconsistencies and noise. Next we evaluate the utility of segment influence (relative to alternative methods) for helping to identify such problematic training data for sequence tagging tasks. 

\subsection{Finding noisy labels in CoNLL}
\label{sec:conll-noisy}

CoNLL has been the \emph{de facto} standard NER dataset in NLP for over a decade, despite known annotation issues \cite{reiss-etal-2020-identifying} including: missing and incomplete entities; incorrect entity labels; questionable entity boundaries, among others. 
To assess whether segment influence can unveil these issues we calculate the influence of train tokens on mispredictions; instances with higher scores are those we would expect to contain labeling errors.

\textbf{Baselines} We evaluate several baselines for identifying noisy instances. At the instance-level, these include instance loss and the $\ell_2$ gradient norm with respect to CRF parameters. For token-level baselines we again compute conditional loss and gradient norms, as well as entropy over the predicted distributions of labels for each token. We aggregate token scores by taking means or maximum values. Higher values suggest noisy annotations.

\begin{figure*}

\centering 
\begin{subfigure}[t]{.329\textwidth}
    \centering 
    \caption{Manually-identified Labeling Errors}
    \includegraphics[width=\textwidth]{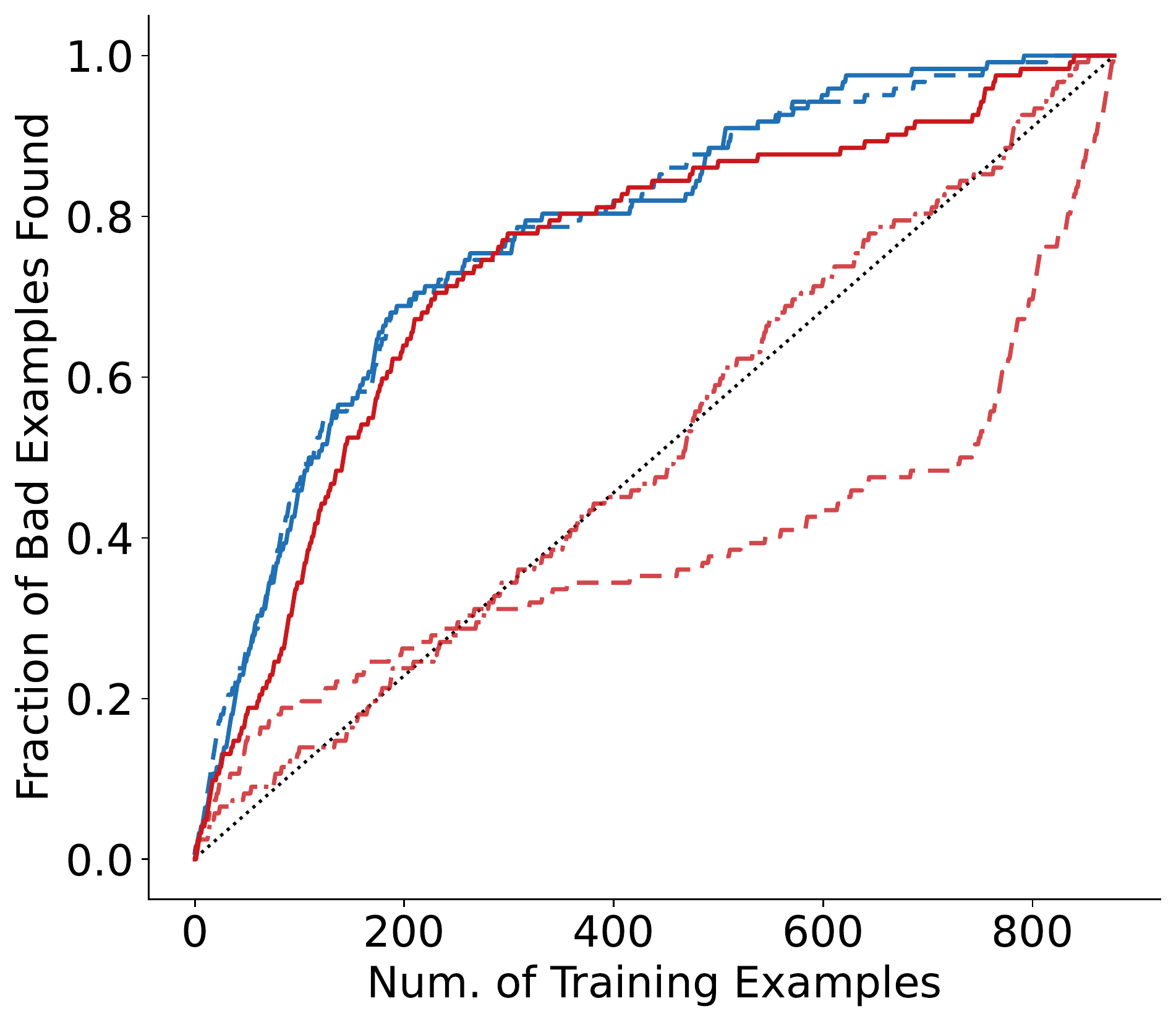}
    
    \label{fig:conll_badness}
\end{subfigure}
\hfill
\begin{subfigure}[t]{.329\textwidth}
    \centering 
    \caption{Random Labeling Errors}
    \includegraphics[width=\textwidth]{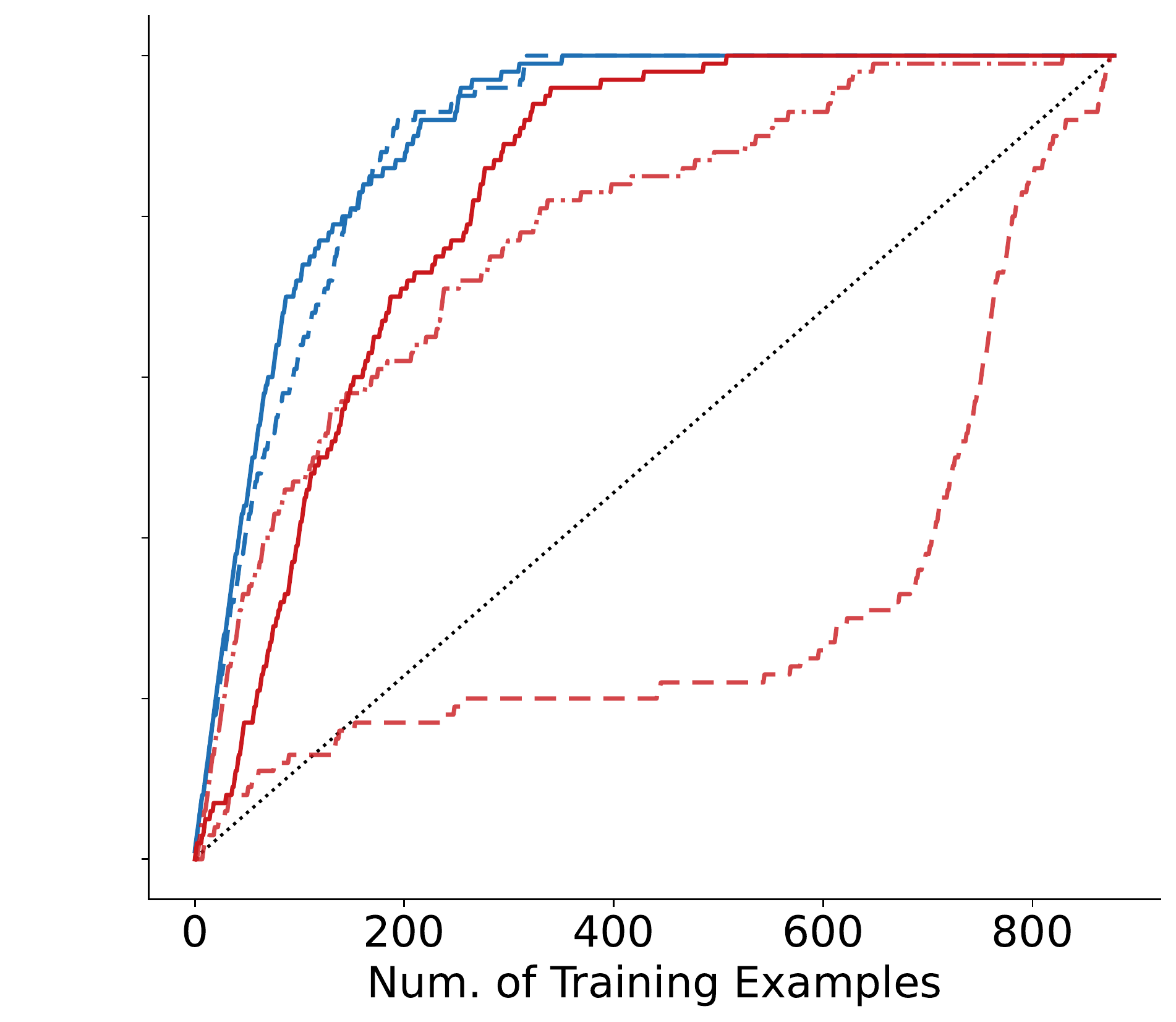}
    
        \label{fig:conll-random-noise}
\end{subfigure}
\hfill
\begin{subfigure}[t]{.329\textwidth}
    \centering 
    \caption{Systematic Labeling Errors}
    \includegraphics[width=\textwidth]{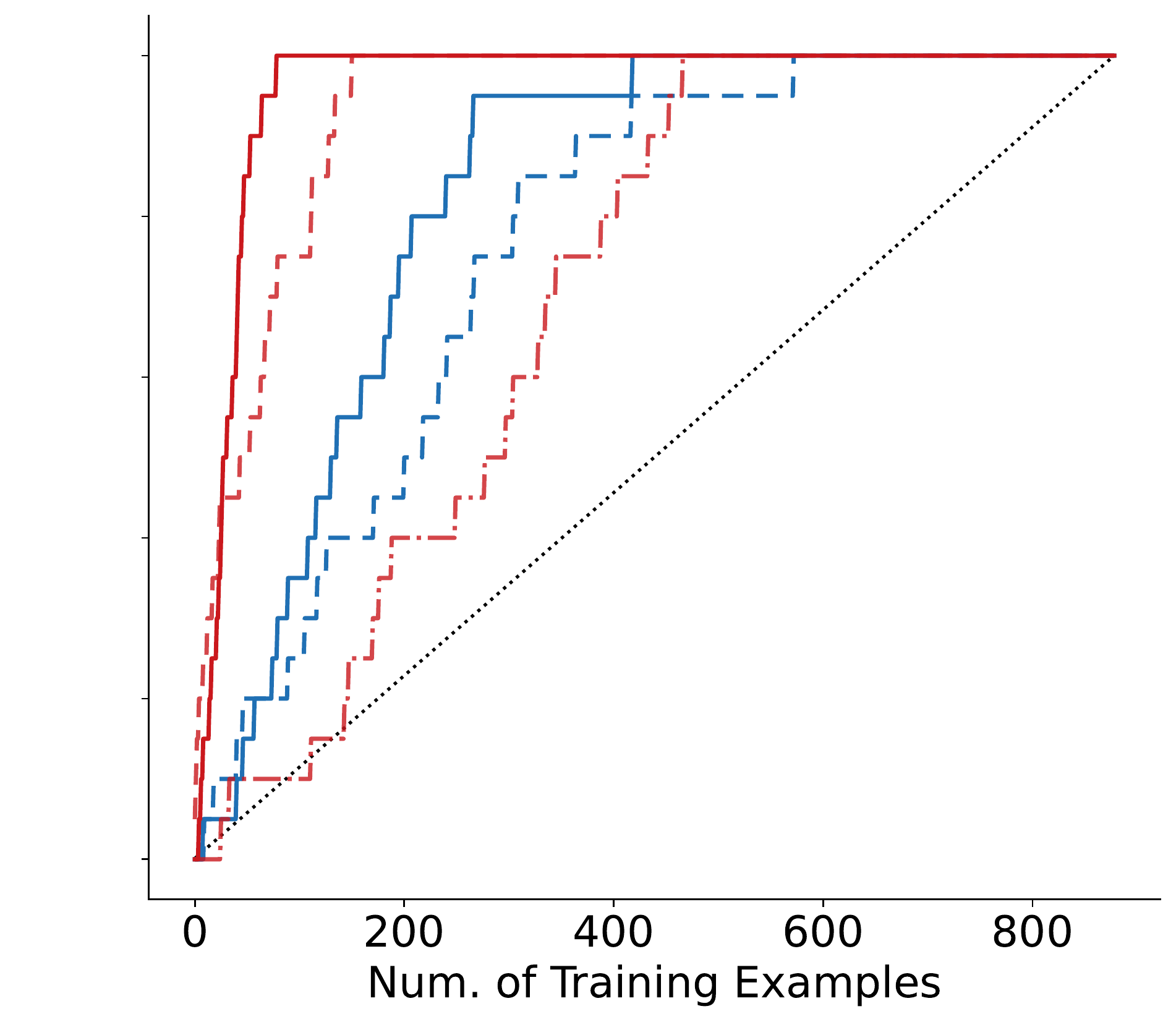}
        \label{fig:conll-sports-noise}
\end{subfigure}
\begin{subfigure}{\textwidth}
\vspace{-1em}
\centering
\includegraphics[width=.65\textwidth]{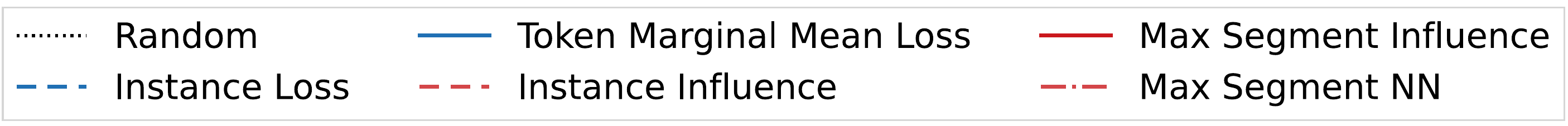}
\end{subfigure}
\caption{Finding problematic CoNLL examples in the train set using different scoring functions. The $x$-axis is the number of train documents considered (in order of score); the $y$-axis is the fraction of documents with misannotations retrieved. }
 \vspace{-.5em}
\end{figure*}

\textbf{Influence-based Methods} We also explore whether a small set of corrected examples (10 instances) can be used together with influence functions to identify noisy examples. The intuition is that if including a training example (or a segment thereof) in the train set increases the loss for a (clean) validation example (or segment), then this training sample may be incorrectly labeled. Therefore, a misannotation score for a training example can then be computed by measuring the influence it appears to exert in the \emph{wrong} direction on the prediction for a clean validation example. Note that a positive influence value indicates that the loss for a validation example will increase if we include the corresponding training example/segment. 

To score training samples using instance-level attribution methods, we average its {\bf instance influence} over the (10) corrected validation samples. To derive a single score using token-level influence, we first compute the {\bf segment influence} of a train token on all validation sample tokens, and then average the top-10 highest values. We take this aggregate measure over only the top-$k$ tokens because averaging over all tokens would lose the granularity captured by segment influence. We also derive a \textbf{segment-level NN} score for a training token by taking dot-product similarities with all validation tokens belonging to a different class and 
averaging the 10 highest values. As above, we compute a training instance misannotation score by taking the mean or maximum of the token level scores.

\vspace{0.35em}
\noindent \textbf{Results} Figure~\ref{fig:conll_badness} plots the rank order of examples under different scoring functions (we include a representative subset of scoring functions here; all results in Appendix~\ref{app:extra-graphs}) against the fraction of documents retrieved which contain any annotation errors identified by \cite{reiss-etal-2020-identifying}. Influence methods efficiently recover noisy training samples, but in this case so do methods such as ranking instances according to average token loss. But this aggregate result may hide differences in the \emph{types} of noise methods are able to identify. We hypothesize that while simpler methods may recover instances with \emph{random} labeling errors, structured influence may better identify \emph{systematic} labeling errors in training data. 

\subsection{Finding Systematic Labeling Errors}

CoNLL contains many apparently \emph{random} annotation errors. These errors lack a discernible pattern and may occur if the annotator makes a mistake due to a lack of focus while annotating. But in addition to these, token labels also exhibit a \emph{systematic inconsistency} where sports teams ({\tt ORG}s) are often mislabeled as {\tt LOC}s. Suppose a practitioner observed a model consistently mislabeling sports teams as locations on held out samples. Ideally, they could recover (and fix) the \emph{source} of this systematically incorrect behavior. By contrast, it is less clear how to reduce \emph{random} mistakes made by the model. We envision the practitioner using segment influence with respect to {\tt ORG} entity tokens mispredicted as {\tt LOC}s to identify (segments of) training points that have resulted in this behavior; these could then be relabeled.

To test whether influence-based methods are better able to unveil systematic labeling errors (compared to simpler methods), we run an experiment using a ``cleaned'' version of CoNLL \cite{reiss-etal-2020-identifying}, which has fixed the  
aforementioned sports team inconsistency.  
We then reintroduce the inconsistency for a subset of training examples about soccer. 
Specifically, we select 20 documents (out of 105) that start with the word `SOCCER' and convert any token labeled as {\tt ORG} within them to {\tt LOC} whenever that token corresponds to a city name in Wikidata (see Appendix~\ref{app:wikidata}). 
We also introduce \emph{random} noise into 100 randomly selected documents in the remaining data, replacing the label for every entity 
with a randomly selected label. 

We apply the scoring methods discussed above to the training set to generate misannotation scores. 
For influence based methods, we randomly select 10 validation documents about soccer. This experiment is set up to simulate conditions where users can interactively identify issues with training data via error analysis. It reflects the envisioned use case, where a practitioner might seek out training data responsible for systematic mispredictions observed on a small held-out set. 
We present results for random and systematic labeling errors in Figures~\ref{fig:conll-random-noise} and ~\ref{fig:conll-sports-noise}, respectively.

For random noise (Figure~\ref{fig:conll-random-noise}), influence-based methods perform slightly worse than our baselines. However, with respect to identifying training samples exhibiting a specific type of noise (Figure~\ref{fig:conll-sports-noise}), influence-based methods substantially outperform baselines. This agrees with the hypothesis that influence-based methods can help efficiently identify noisy examples that have resulted in the model making a specific type of errors. However, in this case instance-level attribution fares almost as well as segment influence. We next report results for a case where segment-level influence offers comparative benefits.

\subsection{Dosage Mislabeling in EBM-NLP}

We consider the task of annotating medical \emph{interventions} (e.g., aspirin) in abstracts of articles describing clinical trials. For this we use the EBM-NLP dataset \cite{nye-etal-2018-corpus}.
Using segment influence, we identified a type of systematic noise in the training set: Annotators (lay workers recruited on Mechanical Turk) sometimes included \emph{dosage} information in their intervention spans, and sometimes did not. This in turn gave rise to apparent errors on the test set, which was labeled by medical experts who did not consider dosages as part of intervention. We provide examples in Table~\ref{tab:dosage-example}.

\begin{table}
\footnotesize
    \centering
    \begin{tabular}{p{.5\columnwidth}|p{.425\columnwidth}}
        Patients received either quality-controlled \tightcbox{blue!15!white}{chloroquine} aiming for a target total dose of \tightcbox{blue!15!white}{\textbf{25 mg base/kg}} in ...
    & \tightcbox{blue!15!white}{dextran-70} infusion was used at the dose of \textbf{7.5 ml/kg} for 30 minutes before CPB ...
    \end{tabular} \vspace{-.5em}
    \caption{Example of dosage mislabeling in the EBM-NLP dataset. In the left instance, dosage is labeled as part of the intervention; in the right instance it is not.} 
     \vspace{-1.25em}
    \label{tab:dosage-example}
\end{table}

\begin{table*}
\scriptsize
    \centering
    \begin{tabular}{l p{4cm} p{4cm} p{4cm}}
    \hline
        {\bf Attribution method} & {\bf Test example} & {\bf Top supporting example} & {\bf Top opposing example} \\ \hline
        Instance Influence & ... assessed the effects of chronic treatment with low doses of \tightcbox{blue!15!white}{aspirin} ( 100 mg/day ) on clinic and ambulatory systolic ( SBP ) ...  & ... the relationship between easy skin bruising and other systemic effects of ICS therapy , including adrenal suppression ... & ... Despite enthusiasm for the use of \tightcbox{blue!15!white}{mild hypothermia} during neurosurgical procedures , this therapy ... \\ \hline
        Segment Influence 
        & \tightcbox{blue!15!white}{Nitrendipine} was given orally at a dosage of 20 \textbf{mg} once ...
        & \tightcbox{blue!15!white}{Alinidine} was administered at a dosage of 30 \textbf{mg} ... &
        The dose of \tightcbox{blue!15!white}{oral famotidine was 2 \textbf{mg/kg/day}} ... 
    \end{tabular}
    \vspace{-0.75em}
    \caption{Dosage label inconsistency as identified via different attribution methods. For each method, we are showing an example of the most typical result achieved. Segment Influence can identify the inconsistent labels for dosage entities in the training dataset. Segment NN retrieves similar examples when explicitly checking for inconsistency.}
     \vspace{-.5em}
    \label{tab:ebm-int}
\end{table*}

We aim to evaluate the degree to which different attribution methods might help practitioners identify such inconsistent labels in the train set. Here we only consider influence based methods: Instance-level influence functions, segment influence functions, and segment nearest neighbors. For each method, we characterize whether the top influential training examples (or segment within) could be used to identify the fact that inconsistent labeling for dosages occur in the training set (leading to the apparent mispredictions on the test set).

Define a \emph{supporting} example as one whose inclusion in training decreases the loss for the test examples/segment with respect to its label (i.e., it has negative influence value), and an \emph{opposing} example as one whose inclusion increases the loss for the same. 
To assess the ability of segment influence to flag systematic dosage mislabeling, we use test examples
in which dosages were mispredicted as interventions, and measure whether: 
(a) the top \emph{supporting} example excludes a dosage from the intervention span, \emph{and} (b) the top \emph{opposing} example has a dosage labeled as intervention. 
Surfacing such conflicting instances as the most influential training points (pointing in opposite directions) for a dosage that appears in a test segment readily suggests the label inconsistency issue at play. 

We find that {\bf instance influence} functions recover such inconsistency in only 35\% of test examples (Table~\ref{tab:ebm-result}), while {\bf segment influence} identifies such cases for $\sim$97\% of these. This supports the use of segment influence for identifying errors in a specific type of entity which may not be apparent when identified via instance influence functions.

\begin{table}
\small
    \centering
    \begin{tabular}{l cc}
    \hline
        Method & Example & Token \\\hline
        Instance inf. & 35.4& --\\
        Segment inf. & 97.9 & 85.4 \\
        Segment NN & 87.5 & 83.3 
    \end{tabular} \vspace{-.5em}
    \caption{Segment influence reliably surfaces dosage labeling inconsistency. For mispredicted dosage test tokens, we report whether top supporting/opposing examples and tokens have conflicting dosage labels.}
     \vspace{-1em}
    \label{tab:ebm-result}
\end{table}

We also consider {\bf segment nearest neighbor}. Here for a given test point where a dosage token has been mispredicted as an intervention, we retrieve the example containing the most similar token (under the token representation induced by $F$) and check if it excludes dosage from its intervention span. 
This occurs for 97\% of the test dosages. However, retrieving the example with the \emph{least} similar token (as an analog to an ``opposing'' example in the case of influence methods) yields no dosages (see  Table~\ref{tab:ebm-int}). This is because dissimilar examples tend not to be useful for analysis. To assess whether segment NN might help verify a hypothesis regarding label inconsistency, we explicitly retrieve the two \emph{most similar} tokens that have the same and a distinct label as the test example, respectively. These two ``nearby'' instances with conflicting labels can be used to check for inconsistency.
Typical examples from each of the above methods are reproduced in Table~\ref{tab:ebm-int}.

\section{Related Work}
\vspace{-.4em}
Influence functions originated in statistics in the context of analyzing linear models \cite{cook1982criticism,chatterjee1986influential,hampel2011robust}. Koh and Liang \citeyearpar{koh2017understanding} reintroduced influence functions to the ML community.

While influence remains the most common method for training data attribution, other methods have also been proposed to identify ``important'' instances, including: Shapley Values \cite{ghorbani2019data}; Fisher Kernels \cite{khanna2019interpreting}; tracking instance gradients during training \cite{pruthi2020estimating,chen2021hydra}; and training surrogate linear models \cite{ilyas2022datamodels}. Another line of work aims to make the computation of influence more efficient \cite{guo2021fastif,51033}. 

Influence functions have been shown to be useful for: aiding training data debugging and artifact identification \cite{han2020explaining,han2021influence,zylberajch-etal-2021-hildif,pezeshkpour-etal-2022-combining}, understanding bias in data \cite{pmlr-v97-brunet19a}, robust optimization \cite{lee2020learning,deng2020interpreting}, active learning \cite{xu2019understanding}, data cleaning \cite{wang2021gradient,kong2022resolving}, and domain adaptation \cite{iter2021trade}. 

Some work has also 
combined influence functions with \emph{feature attribution} (e.g., integrated gradients) to point to specific tokens within instances that were influential \cite{koh2017understanding,pezeshkpour-etal-2022-combining,zhang-etal-2021-sample}. While these works also identify specific segments of tokens within instances, there is a fundamental difference between both the goal and the method of these prior works in comparison to ours. These prior efforts provided mechanisms to approximate which tokens most contribute to the influence score of a given example (that is, locating parts of the \emph{input} text which, if perturbed, would greatly affect the influence of said point). They essentially compute the gradient of influence values for an \emph{instance-level} prediction with respect to the input tokens. By contrast, we provide a machinery to measure influence for structured, \emph{multi-part} outputs. Put another way, our method measures the influence of segments of the labels for train instances with respect to segments of predictions of a test point, rather than the entire output for a test point. 

Work on interpretability for \emph{structured prediction} tasks have mostly focused on feature attribution for token-level predictions. 
Such methods have been used to characterize model behavior \cite{agarwal-etal-2021-interpretability,alvarez-melis-jaakkola-2017-causal}, extract rationales \cite{vafa2021rationales}, analyze internal model representations \cite{clark2019does,vig2020bertology} and debug erroneous text generations \cite{strobelt-etal-2018-debugging}. Although rare, some structured prediction works have attempted to trace model behavior back to the training set but they invariably use influence with respect to whole structured output \cite{wang2021gradient,51033}. 

\section{Conclusions}
\vspace{-.4em}
We have presented a method for computing token-level influence values for sequence tagging tasks, i.e., \emph{segment influence}. 
We validated this method via synthetic experiments, ascertaining that it retrieves the expected tokens as `influential' for predictions made on held-out examples containing specific artifacts. 
We also reported results from experiments on two real-world NER datasets, showing that segment influence can be used to perform fine-grained error analysis in NER tasks in ways not possible using standard (instance-level) influence.

\section{Limitations}
\label{sec:Limitations and Generalization}

In this paper we provided a method to compute influence functions for token level predictions in sequence tagging tasks, and showed the utility of this with respect to identifying noisy labels in NER training data. 
The exact influence functions defined in Section~\ref{sec:exact-inf} do not explicitly depend on the sequential nature of the output or the use of a Linear Chain CRF structure; the same derivation can be used for any structured prediction model. However, there are important limitations to this method, both theoretical and in terms of efficiency.

The immediate problem occurs in the derivation of Exact Influence. The objective for retraining after removal of a segment is not convex due to the presence of marginal likelihood term (Equation~\ref{eq:exact-inf-retrain}) and therefore susceptible to presence of multiple minima. 

We have assumed a probabilistic model over outputs conditioned on an input sequence, and that the training objective is to minimize the negative log-likelihood over the training data under this model. 
The approach may not be readily amenable to alternative loss functions or model classes (e.g., structured SVMs; \citealt{tsochantaridis2004support}).

Even considering only graphical models for structured prediction, our approach requires the ability to compute conditional probabilities for any subset of outputs and the corresponding gradient for model parameters.
We have focused on linear-chain CRFs, which permits simple computation of conditional probabilities and gradients; this may not be the case for other such structured prediction tasks (See Appendix~\ref{app:cond-gen}). 

Finally, we reiterate the efficiency issues from Section~\ref{sec:challenges} prevalent for sequence tagging tasks themselves. 
Even with simplifications (using only the final layer parameters and ignoring the hessian matrix), we must store a vector of size at least $d \cdot C$ for every segment within every example in the training set ($d$ denoting the feature vector size and $C$ the number of classes). 

As an example, consider CoNLL dataset trained using BERT-base model $(d=768, C=9)$. 
A single vector requires 27KB of space; storing the vector for every segment in every training example requires $\sim$240GB of space. Clearly, this is infeasible for larger datasets. Even restricting to token level (as we have above) requires 9GB of space. To scale up to larger datasets and sequence lengths, we need a way to reduce these space requirements, perhaps via compression. In addition, at token level, we may reduce the size of vector needed to be stored to $(d + C)$ by exploiting the fact that the gradient can be written as a outer product of the feature and the error vector (See Appendix~\ref{app:crf-grad}); in CoNLL, for example, this can reduce space requirements by 9x. 

\section{Broader Impact Statement}
\label{sec:ethics}

Large-scale pre-trained models for NLP are being deployed with increasing frequency, given their empirical success. 
But these models remain opaque, 
making it difficult to 
know why a model made any specific prediction. 
Training data problems such as artifacts, biases, and labeling errors constitute common sources of model misbehaviors, and instance attribution methods---such as what we have proposed in this work---provide a potential mechanism to unearth and ultimately fix such issues.
However, while influence methods like the proposed segment influence functions may provide one means to identify training data issues, we would caution against using absence of errors discovered via these methods as the proof of their non-existence in the dataset. 
These tools are not perfect, and one should make an independent judgment of the benefits and risks of releasing a newly developed models whose behavior we may not \emph{fully} understand. 

We also note that segment influence methods require significant storage and compute resources, with concomitant environmental implications. 
Practitioners should therefore weigh the potential benefits of using these methods as error analysis tools against the energy consumption costs. 
We hope that future developments in efficient storage and computation of influence vectors might soon mitigate this particular consideration. 

\bibliography{main}
\bibliographystyle{acl_natbib}

\appendix 

\section{Derivation of Influence Functions}
\subsection{Instance Influence Functions}
\label{app:instance-inf-deriv}

\newcommand{\ntheta}{\hat{\theta}_{\epsilon,i}}
\newcommand{\dtheta}{\delta_\epsilon}

We provide a derivation of instance Influence Functions here for completeness (originally derived by \citealt{koh2017understanding}). 
Consider $\htheta$, the parameters that minimize the loss function $\mathcal{L}$ over our training dataset.

\begin{equation}
    \joint(\mathcal{D}, \theta) = \frac{1}{N} \sum_{i=1}^N \joint(z_i, \theta)
\end{equation}

\noindent The loss function is assumed to be twice-differentiable and strongly convex, such that a positive-definite Hessian $\nabla^2_{\theta} \joint(\mathcal{D}, \theta)$ exists. 

Influence Functions measure the change in the loss of some test example $z_t$ if we slightly upweigh the training example $z_i$ during training. Under such upweighting, the new parameters can be written as:

\begin{equation}
\label{eq:new-ins=inf-loss}
    \ntheta = \underset{\theta}{\arg\!\min}\hspace{.2em} \joint(\mathcal{D}, \theta) + \epsilon \joint(z_i, \theta)
\end{equation}

Define the change in parameters as $\dtheta = \ntheta - \htheta$.
The rate of change of parameters with respect to $\epsilon$ can be written as:

\begin{equation}
\label{eq:change-eq-deriv}
    \frac{d\dtheta}{d\epsilon} = \frac{d\hat{\theta}_{\epsilon,i}}{d\epsilon}
\end{equation}

\noindent Since the new parameters minimize the perturbed loss function, we can consider its first-order optimality criteria:

\begin{equation}
    \grad \joint(\mathcal{D}, \ntheta) + \epsilon \grad \joint(z_i, \ntheta) = 0
\end{equation}

\noindent Assuming $\ntheta \rightarrow \htheta$ as $\epsilon \rightarrow 0$, we perform a Taylor-expansion from the left hand side around $\htheta$:\footnote{Here, $\nabla^2_{\theta} \joint(\mathcal{D}, \htheta)$ is the hessian of the loss $\joint(\mathcal{D}, \htheta)$}

\begin{align*}
    \grad \joint(\mathcal{D}, \htheta) &+ \epsilon \grad \joint(z_i, \htheta) \\
    &+ [\nabla^2_{\theta} \joint(\mathcal{D}, \htheta) + \epsilon \nabla^2_{\theta} \joint(z_i, \htheta)] \dtheta = 0
\end{align*}

\noindent Dropping the higher order terms in $|\dtheta|$ and noting that the first term is zero (due to first-order optimality of original loss; $\grad\joint(\mathcal{D}, \htheta) = 0$), we arrive at the following equation:

\begin{align*}
    &\epsilon \grad \joint(z_i, \htheta) + [\nabla^2_{\theta} \joint(\mathcal{D}, \htheta) + \epsilon \grad \joint(z_i, \htheta)] \dtheta = 0 \\
    &\dtheta = - [\nabla^2_{\theta} \joint(\mathcal{D}, \htheta) + \epsilon \nabla^2_{\theta} \joint(z_i, \htheta)]^{-1} \epsilon \grad \joint(z_i, \htheta) \\
    &\dtheta = - [\nabla^2_{\theta} \joint(\mathcal{D}, \htheta)]^{-1} \epsilon \grad \joint(z_i, \htheta)
\end{align*}

\noindent Where in the last step we have dropped the higher order terms in $\epsilon$. 
Using above expression and Equation~\ref{eq:change-eq-deriv}, we get the final form of the derivative of the parameters with respect to $\epsilon$:

\begin{equation}
    \frac{d\hat{\theta}_{\epsilon,i}}{d\epsilon} = - [\nabla^2_{\theta} \joint(\mathcal{D}, \htheta)]^{-1} \grad \joint(z_i, \htheta)
\end{equation}

Given the equation for derivative of parameters, we can compute the derivative of the loss of the test example with respect to $\epsilon$ by chain rule:

\begin{equation}
\begin{split}
    &\frac{d \joint(z_t, \htheta_{\epsilon,i})}{d\epsilon} = \frac{d \joint(z_t, \htheta)}{d\theta}^{\top} \frac{d\hat{\theta}_{\epsilon,i}}{d\epsilon} \\
    &=- \grad\joint(z_t, \htheta) [\nabla^2_{\theta} \joint(\mathcal{D}, \htheta)]^{-1} \grad \joint(z_i, \htheta)
\end{split}
\end{equation}

\subsection{Segment Influence Functions}
\label{app:segment-inf-deriv}

The derivation for Segment Influence Functions is similar to the proceeding. 
The only difference involves replacing the $\joint(z_i, \theta)$ in Equation~\ref{eq:new-ins=inf-loss} with the conditional loss of the segment $\conditional(z_i^{[a, b]}, \theta)$, yielding: 

\begin{equation}
    \hat{\theta}_{\epsilon,i[a,b]} = \underset{\theta}{\arg\!\min}\hspace{.2em} \joint(\mathcal{D}, \theta) + \epsilon \hspace{.2em} \conditional(z_i^{[a,b]}, \theta)
\end{equation}

\noindent The rest of the derivation remains the same, and ultimately provides the derivative of the parameters with respect to $\epsilon$ as:

\begin{equation}
     \frac{d\hat{\theta}_{\epsilon,i[a,b]}}{d\epsilon} = - [\nabla^2_{\theta} \joint(\mathcal{D}, \htheta)]^{-1} \grad \conditional(z_i^{[a,b]}, \theta)
\end{equation}

\section{Gradient with Respect to CRF Parameters}
\label{app:crf-grad}

\newcommand{\sumab}{\sum_{y'_a \in \mathcal{Y}}\dots\sum_{y'_b \in \mathcal{Y}}}
\newcommand{\W}[1]{\vec{y}_{#1}^\top \mathbold{W} F(x)_{#1}}
\newcommand{\Ts}[2]{\vec{y}_{#1}^\top \mathbold{T} \vec{y}_{#2}}
\newcommand{\Wp}[1]{\vec{y'}_{#1}^\top \mathbold{W} F(x)_{#1}}
\newcommand{\Tsp}[2]{\vec{y'}_{#1}^\top \mathbold{T} \vec{y'}_{#2}}

We begin by reiterating the definition of joint probability under a CRF of an instance $(x, y)$ and the marginal probability of a partial label sequence $y_{-[a,b]}$ below:

\begin{align*}
    p(y|x) &= \frac{e^{s(y, x)}}{Z(x)} \\
    p(y_{-[a,b]} | x) &= \sumab \frac{e^{s(y', x)}}{Z(x)} \\
\end{align*}

\noindent Where $Z(x)$ is the normalization term for the CRF, which is independent of sequence labels (i.e., depends only on $x$) and $y' = y_{-[a,b]} \cup \{y'_a, \dots, y'_b\}$. 

Using above definitions, the conditional probability under a CRF of a segment $[a, b]$ for a instance $(x, y)$ is given by:

\begin{equation}
\begin{split}
    p(y_{[a,b]}&|y_{-[a,b]}, x) = \frac{p(y | x)}{p(y_{-[a,b]} | x)} \\
        &= \frac{e^{s(y, x)}}{\sumab e^{s(y', x)}}
\end{split}
\end{equation}

In a linear-chain CRF, the score function $s(y, x)$ can be divided into sum of three parts. 
Therefore we can write $e^{s(y, x)}$ as a product of three parts:

\begin{itemize}
    \item Terms depending only on $y_{-[a, b]}$: These are terms of form $\W{t}$ and $\Ts{t-1}{t}$ where $t, t - 1 \notin [a, b]$. Call them collectively as $s_{-[a, b]}$
    \item Terms depending only on $y_{[a, b]}$: These are terms of form $\W{t}$ and $\Ts{t-1}{t}$ where $t, t - 1 \in [a, b]$. Call them collectively as $s_{[a, b]}$.
    \item Interaction Terms: Only two such terms exist -- $\Ts{a-1}{a}$ and $\Ts{b}{b+1}$. Call them collectively as $T_I$.
\end{itemize}

\noindent (As in the main text, we use $\vec{y}_{t}$ to denote the one-hot representation of label $y_{t}$.)
Applying it to the formula for conditional probability above, we have:

\begin{align*}
    p(y_{[a,b]} &|y_{-[a,b]}, x) \\
    &= \frac{e^{s_{-[a, b]}}e^{s_{[a, b]} + T_I}}{\sumab e^{s_{-[a, b]}}e^{s_{[a, b]} + T_I}} \\
    &= \frac{e^{s_{-[a, b]}}e^{s_{[a, b]} + T_I}}{e^{s_{-[a, b]}} \sumab e^{s_{[a, b]} + T_I}} \\
    &= \frac{e^{s_{[a, b]} + T_I}}{\sumab e^{s_{[a, b]} + T_I}}
\end{align*}

In Step 2 above, we note that the terms collectively represented by $s_{-[a, b]}$ do not depend on any of summation variables, so we can take them out of the summation in the denominator.

The formula for conditional probability, as achieved, is similar to standard CRF probability and therefore, the gradient of its (negative-) logarithm (i.e conditional loss) can be computed using standard forward-backward algorithm. At token level, the expression for conditional probability can be simplified even further:

\begin{align*}
    p(y_t &|y_{-t}, x) \\ 
    &= \frac{e^{\Ts{t-1}{t} + \W{t} + \Ts{t}{t+1}}}{\sum_{y'_t \in \mathcal{Y}} e^{\Tsp{t-1}{t} + \Wp{t} + \Tsp{t}{t+1}}} 
\end{align*}
\begin{align*}
    \log p&(y_t |y_{-t}, x) \\
    &= \Ts{t-1}{t} + \W{t} + \Ts{t}{t+1} \\
    &- \log \sum_{y'_t \in \mathcal{Y}} e^{\Tsp{t-1}{t} + \Wp{t} + \Tsp{t}{t+1}}
\end{align*}

We can now take the gradient with respect to both emission parameters ($\mathbold{W}$) and transition parameters ($\mathbold{T}$). In the following equations, the $\otimes$ indicate the outer product of two vectors. In addition, we use the fact that gradient of a bilinear form $\vec{u}^{\top} \mathbold{A} \vec{v}$ with respect to a matrix $\mathbold{A}$ is $\vec{u} \otimes \vec{v}$.

\begin{align*}
    \nabla_\mathbold{W} &\log p(y_t |y_{-t}, x) \\ 
    &= \vec{y}_t \otimes F(x)_t - \sum_{y'_t \in \mathcal{Y}} p(y'_t |y_{-t}, x) \vec{y}'_t \otimes F(x)_t \\
    &= (\vec{y}_t - \sum_{y'_t \in \mathcal{Y}} p(y'_t |y_{-t}, x) \vec{y}'_t) \otimes F(x)_t \\
    &= \vec{e}_t \otimes F(x)_t \\
    \nabla_\mathbold{T} &\log p(y_t |y_{-t}, x) \\
    &= [\vec{y}_{t-1} \otimes \vec{y}_t + \vec{y}_t \otimes \vec{y}_{t+1}] \\
    &- \sum_{y'_t \in \mathcal{Y}} p(y'_t |y_{-t}, x) [\vec{y}_{t-1} \otimes \vec{y}'_t + \vec{y}'_t \otimes \vec{y}_{t+1}] \\
    &= \vec{y}_{t-1} \otimes (\vec{y}_t - \sum_{y'_t \in \mathcal{Y}} p(y'_t |y_{-t}, x) \vec{y}'_t) \\
    &+ (\vec{y}_t - \sum_{y'_t \in \mathcal{Y}} p(y'_t |y_{-t}, x) \vec{y}'_t) \otimes \vec{y}_{t+1} \\
    &= \vec{y}_{t-1} \otimes \vec{e}_t + \vec{e}_t \otimes \vec{y}_{t+1}
\end{align*}

Note that the gradient with respect to $\mathbold{W}$ decomposes as outer product of an error vector $\vec{e}_t = (\vec{y}_t - \sum_{y'_t \in \mathcal{Y}} p(y'_t |y_{-t}, x) \vec{y}'_t)$ and feature vector $F(x)_t$. This helps us store the feature vector and error vector separately, taking the space $O(d + C)$, rather than storing the computed outer product with space complexity $O(d \cdot C)$.

\section{Modeling Details}
\label{app:data-model}

The models for both CoNLL (with additional noise where required) and EBM-NLP were trained using the PyTorch v1.10.1 ~\cite{NEURIPS2019_9015} library and {\tt BigBird-base} (123M parameters) as default transformer encoders, as implemented in HuggingFace v4.12.5~\cite{wolf-etal-2020-transformers}. 
We used token encodings from the last model layer and fed these through a CRF model with parameters $W$ and $T$. 
Since transformer based models return feature vectors for wordpieces, we obtain the feature vector for the word by taking an average over the constituent wordpieces within a word. 
We followed this with a a dropout layer (with probability 0.3).

We trained models using the Adam Optimizer \cite{kingma2014adam} for 15 epochs with a learning rate of 2e-5. We used gradient clipping with a maximum value of 10; best model checkpoint (for evaluation) were selected on the basis of validation set losses. 

For both CoNLL and EBM-NLP, we achieve similar performance as reported in previous works. For CoNLL, we achieve 93.6\% Exact-match F1 score on the validation set (and 92\% on the test set) in Section~\ref{sec:conll-noisy}, comparable to the best performance at a constructed benchmark of CoNLL at PapersWithCode\footnote{\url{https://tinyurl.com/2p95ymav}}.

For EBM-NLP, we achieve 73.3\% token-level F1 score on the test set (and 72.5\% on the validation set), which is comparable to current best performance achieved by PubmedBERT~\cite{gu2021domain}.  

All experiments were run using single Nvidia v100 GPU. It took 45 minutes to train a single CoNLL model and 90 minutes for EBM-NLP.

\subsection{Syntactic features for the Linear Model}
\label{app:syntac-feats}
For the linear model trained in Section~\ref{sec:corr}, we use the following features for each token (in addition to their corresponding GloVe embeddings): parts-of-speech tags; indicators for digits, capitalization, title-casing, and ``stop-words''. 
We also include this information for the tokens immediately (left and right) adjacent. 
These features are derived by passing CoNLL documents through {\tt SpaCy v3.0.7} \cite{Honnibal_spaCy_Industrial-strength_Natural_2020}.

\section{WikiData Gazetteer}
\label{app:wikidata}

To identify city names in CoNLL dataset for soccer experiments, we use the Neckar tool \cite{10.1007/978-3-319-73706-5_10}, which extracts WikiData entities and associates with them a label in the set {\tt PER}, {\tt LOC} and {\tt ORG}. 
We use the version 1.0 dump of entities and extract all the entities labeled as {\tt LOC}, along with their WikiData aliases. 
We perform post-processing to remove common mislabelings of words as {\tt LOC} (words that indicate a month or a day of week) and use the final set of entity names to identify locations in CoNLL using (lowercase normalized) exact string match.

\section{CoNLL Labeling Errors: All Comparisons}
\label{app:extra-graphs}

\begin{figure*}
\begin{subfigure}{.45\textwidth}
\caption{Manually-annotated Labeling Errors}
    \centering
    \includegraphics[width=\textwidth]{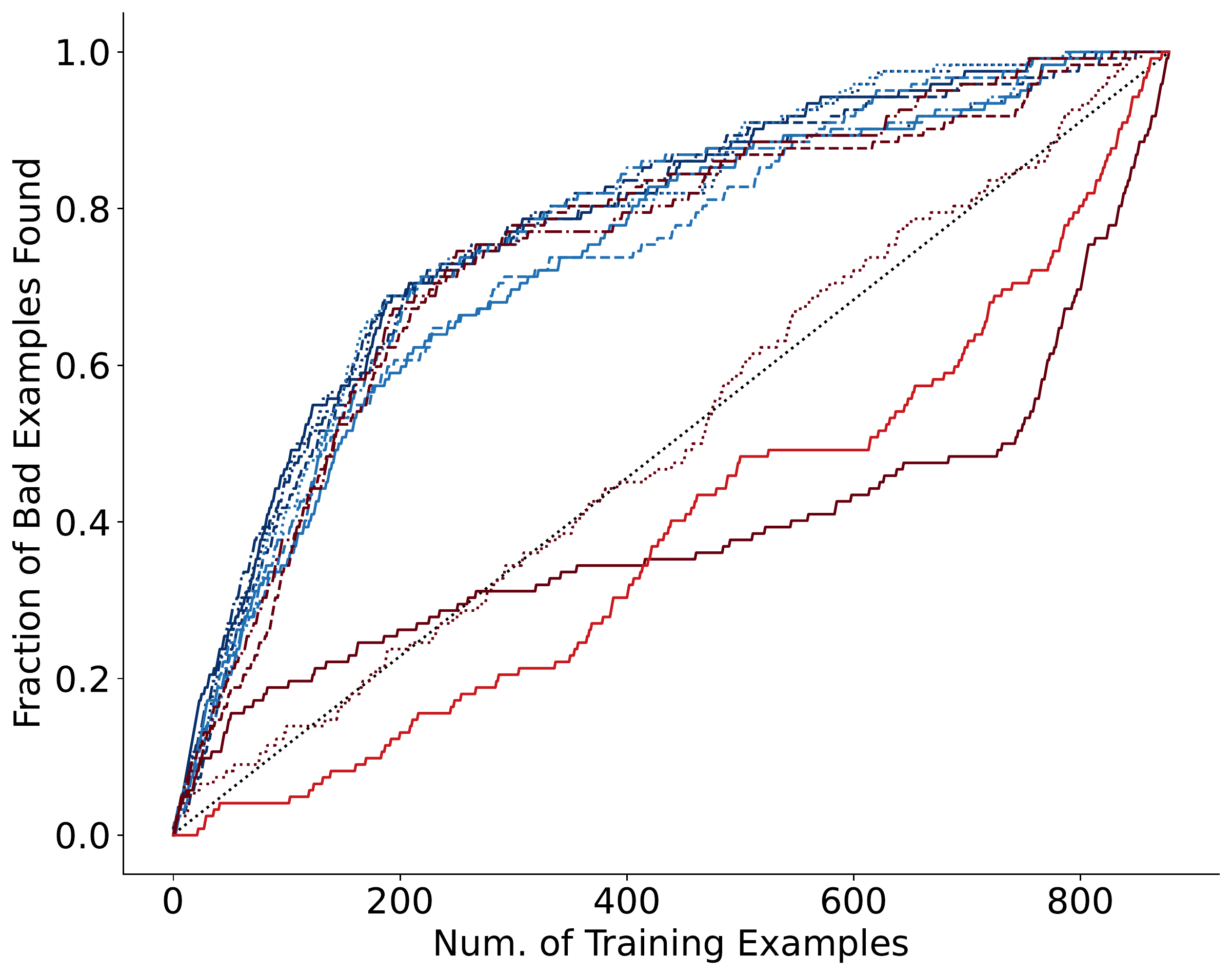}
    \label{fig:conll-all-graphs-errors}
\end{subfigure}
\begin{subfigure}{.45\textwidth}
\caption{Random Labeling Errors}
    \centering
    \includegraphics[width=\textwidth]{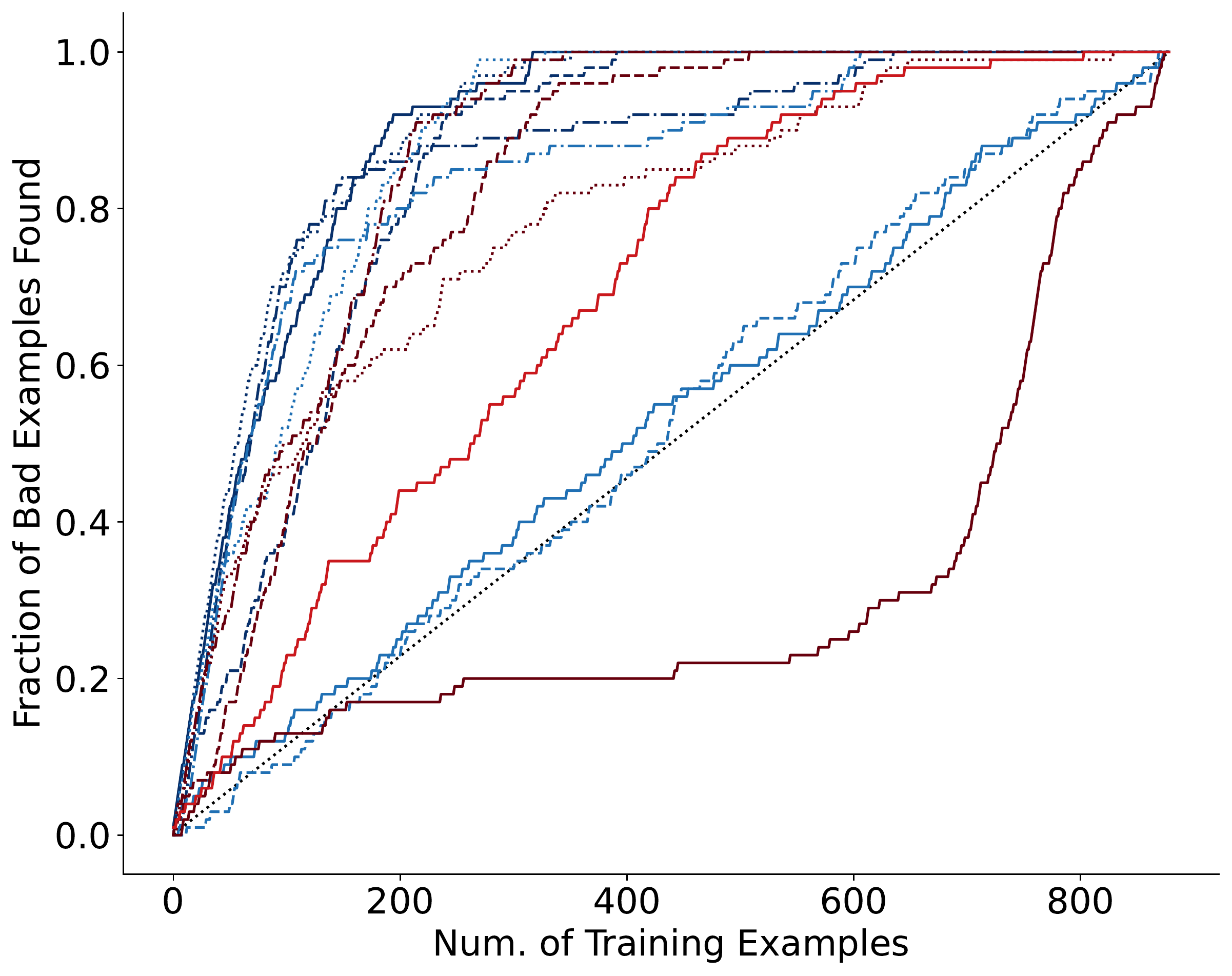}
\label{fig:conll-all-graphs-random}
\end{subfigure}
\begin{subfigure}{.45\textwidth}
\caption{Systematic Labeling Errors}
    \centering
    \includegraphics[width=\textwidth]{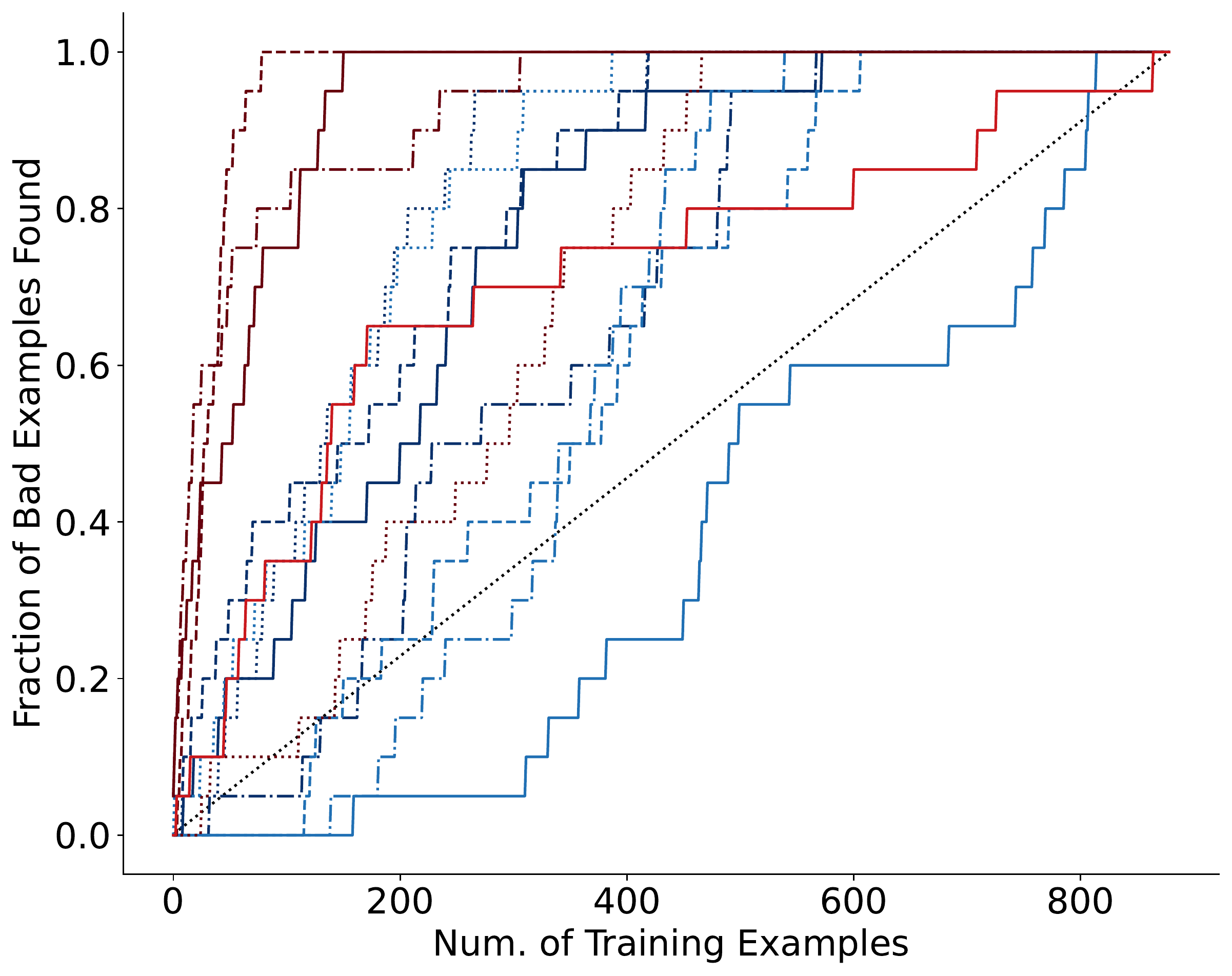}
\label{fig:conll-all-graphs-org-loc}
\end{subfigure}
\begin{subfigure}{.4\textwidth}
    \centering
    \includegraphics[width=\textwidth]{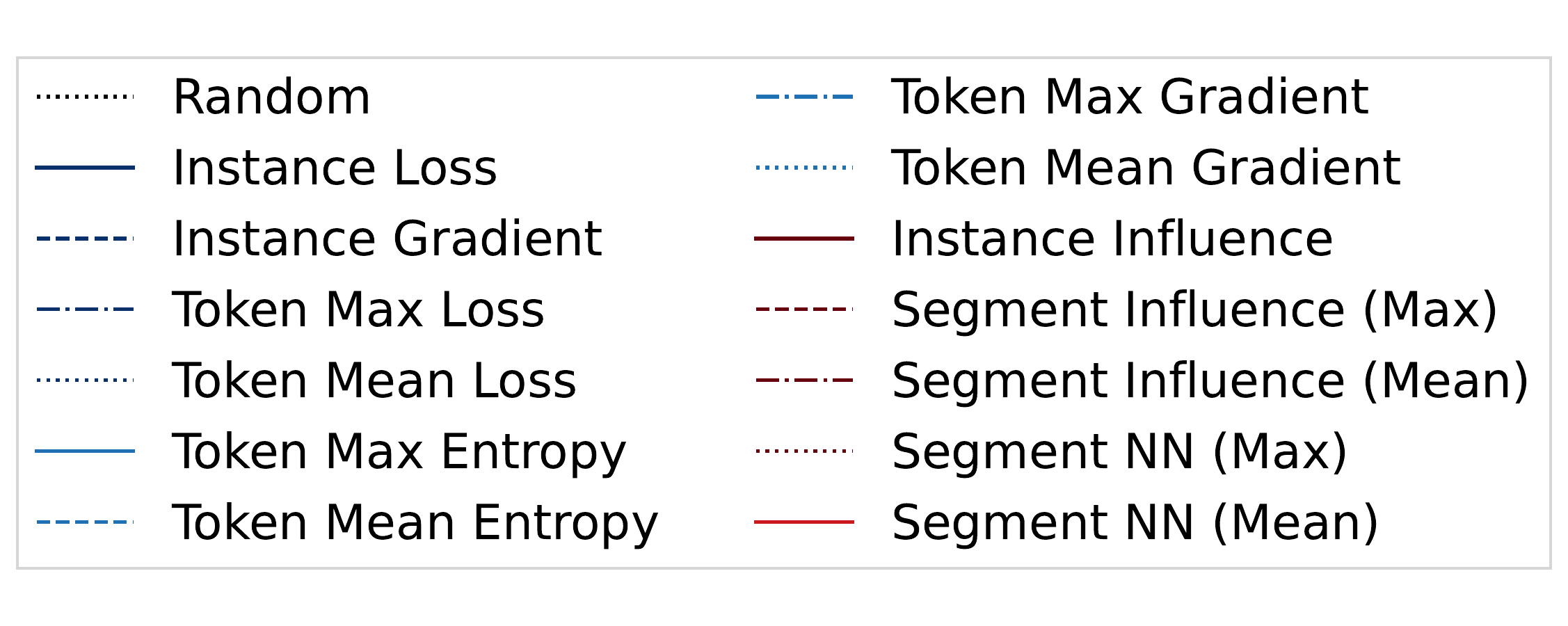}
\end{subfigure}
\vspace{-.5em}
\caption{Finding problematic CoNLL examples in the train set using different scoring functions. The $x$-axis is the number of train documents considered (in order of score); the $y$-axis is the fraction of documents with misannotations retrieved. }
\end{figure*}

In Figure~\ref{fig:conll-all-graphs-errors}, \ref{fig:conll-all-graphs-random} and \ref{fig:conll-all-graphs-org-loc}, we present the respective counterparts to Figure~\ref{fig:conll_badness} showing the performance of all baselines and influence based methods for identifying examples with labeling errors in CoNLL dataset.

\section{Dosage Regex}
\label{app:dose-regex}

To identify dosages in EBM-NLP, we apply the following regex to sentences and identify non-overlapping matches. 
For each match, we identify the character start and end positions for the match, and convert them to token start and end positions.

\begin{lstlisting}
@STRENGTH_UNIT@::mg/dl|mg/ml|g/l|milligrams|milligram|mg|grams|gram|g|micrograms|microgram|mcg|meq|iu|cc|units|unit|tablespoons|tablespoon|teaspoons|teaspoon|mg/kg|IU
@STR_NUM@::half|one|two|three|four|five|six|seven|eight|nine|ten|twelve
@DECIMAL_NUM@::(?:\\d+,)?\\d+(?:\\.\\d+)?(?:(?: |-)?(?:-|to)(?: |-)?(?:\\d+,)?\\d+(?:\\.\\d+)?)?

strength::\b(@DECIMAL_NUM@/)?(@DECIMAL_NUM@)(\s+|-)?(@STRENGTH_UNIT@)\b
strength::\b(@DECIMAL_NUM@/)?(@DECIMAL_NUM@)\s?%
strength::\b\d+\s?(-|to)\s?\d+(\s|-)?(@STRENGTH_UNIT@)\b
strength::\b(@STR_NUM@)\s+(to\s+(@STR_NUM@)\s+)?(@STRENGTH_UNIT@)\b
\end{lstlisting}

\section{EBM-NLP Results}

Table~\ref{tab:ebm-result-full} provides the complete counterpart to Table~\ref{tab:ebm-result}.

\section{Future Work: Conditional Generation}
\label{app:cond-gen}
We conclude by considering the case of identifying influential training data in sequence-to-sequence prediction tasks using the proposed method. 
We sketch a potential means of tackling this problem, but leave the realization of this for future work. 
Typically one defines an autoregressive model of output tokens in such problems: 

\begin{equation}
    p_{\theta}(y_1, \dots, y_L|X) = \prod_{t=1}^L p_{\theta}(y_t | y_1, \dots, y_{t-1})
\end{equation}

\noindent The conditional probability of any single output token is given by:

\begin{equation}
\begin{split}
    \pth(y_t | y_1, \dots, y_{t-1}, y_{t+1}, \dots, y_L, X) = \\
    \frac{\pth(y_1, \dots, y_L|X)}{\sum_{v=1}^{|\mathcal{V}|} \pth(y_1, \dots, y_t = v, \dots, y_L|X)}
\end{split}
\end{equation}

Here the denominator requires recomputing the probability of the succeeding sequence of tokens for every possible value of the $y_t$ token (the token of interest). 
This may be infeasible when the vocabulary size is large. 
Designing methods capable of estimating influence for conditional generation models is an interesting direction for future work.

\begin{table*}
\small
    \centering
    \begin{tabular}{l|ccc|ccc}
        
        Method & Supporting & Opposing & Conflicting & Supporting & Opposing & Conflicting \\
         & example & example & & token & token & \\\hline
        Instance inf. & 68.7 & 58.3 & 35.4 & -- & -- & --\\
        Segment inf. & {\bf 97.9} & {\bf 1.00} & {\bf 97.9} & {\bf 97.9} & \textbf{87.5} & \textbf{85.4}\\
        Segment NN & {\bf 97.9} & {\bf 89.5} & {\bf 87.5} & \textbf{97.9} & 85.4 & 83.3 \\
    \end{tabular}
    \caption{Segment influence reliably surfaces inconsistency in terms of dosage labeling. For every mispredicted dosage token in the test set, we evaluate whether: (1) The top \textbf{supporting example} contains a dosage token outside of a marked intervention span; (2) The top \textbf{opposing example} includes an intervention span containing a dosage token, and; (3) whether both of these conditions are true, i.e., whether the top supporting/opposing examples are conflicting. We report percentages for each.} 
    \label{tab:ebm-result-full}
\end{table*}

\end{document}